\pdfoutput=1
\documentclass[10pt,twocolumn,letterpaper]{article}
\ifdefined\pdfinfoomitdate\pdfinfoomitdate=1 \pdftrailerid{}\pdfsuppressptexinfo=-1 \fi

\usepackage[pagenumbers]{cvpr}

\usepackage{array}
\usepackage{pifont}

\definecolor{linkblue}{rgb}{0.21,0.49,0.74}
\usepackage[pagebackref,breaklinks,colorlinks,allcolors=linkblue]{hyperref}

\title{RoboTeam: Benchmarking Vision-Based Multi-Agent Control\\of Many Robot Arms in Real Time}

\begin{document}
\setcounter{topnumber}{1}
\twocolumn[{%
\renewcommand\twocolumn[1][]{#1}%
\begin{center}
{\Large\bf \thetitle\par}
\vspace{5pt}
{\fontsize{11}{13.5}\selectfont Changxuan~Fan\textsuperscript{1}, Zhongwei~Xie\textsuperscript{1}, Boyi~Zhang\textsuperscript{1}, Shaoli~Hu\textsuperscript{1}, Chang~Xu\textsuperscript{2}, Xiyu~Ren\textsuperscript{1}, Yufei~Li\textsuperscript{1},\\
Qiao~Xiao\textsuperscript{1}, Huihao~Jing\textsuperscript{1}, Kawai~Chung\textsuperscript{1}, Wenbin~Hu\textsuperscript{1}, Hongyu~Luo\textsuperscript{1}, Lingyun~Xie\textsuperscript{1}, Yisen~Gao\textsuperscript{1},\\
Chunyang~Li\textsuperscript{1}, Haoyu~Huang\textsuperscript{2}, Wuganjing~Song\textsuperscript{1}, Yuxuan~Liu\textsuperscript{1}, Chiu~Yeung~Tang\textsuperscript{1}, Fangneng~Zhan\textsuperscript{1}, Xi~Yang\textsuperscript{1}\par}
\vspace{2pt}
{\small \textsuperscript{1}The Hong Kong University of Science and Technology\quad \textsuperscript{2}Microsoft\qquad{\footnotesize\tt cfanam@connect.ust.hk}\par}
\vspace{1pt}
{\small Project page: \href{https://changxuanfan.github.io/RoboTeam/}{\texttt{changxuanfan.github.io/RoboTeam}}\qquad Code: \href{https://github.com/changxuanfan/RoboTeam}{\texttt{github.com/changxuanfan/RoboTeam}}\par}
\end{center}
\vspace{2pt}
\begin{center}
  \centering
  \captionsetup{type=figure}
  \includegraphics[width=\linewidth]{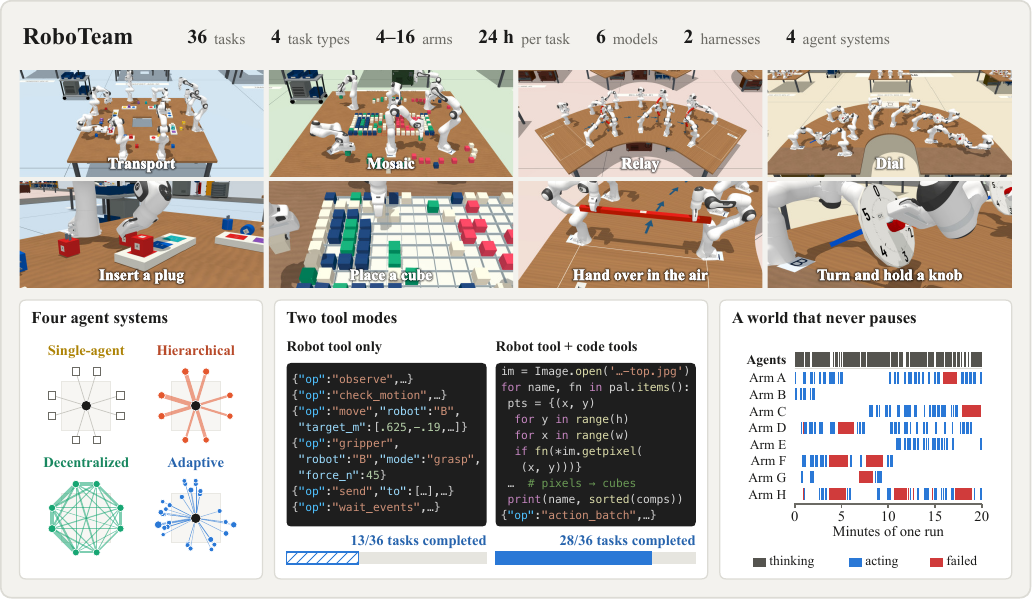}
  \caption{RoboTeam at a glance. Top: the four task types and one step of each. Bottom: the four agent systems (dots are agents, squares are arms), the two tool modes with real calls, and 20 minutes of one run.}
  \label{fig:overview}
\end{center}
}]
\begin{abstract}
Multimodal agents can now control robot arms by calling tools and writing code.
Existing benchmarks fall short in three ways: most test one or two arms on short tasks, many pause the world while the agent thinks, and many give it ground-truth object positions.
We present \textbf{RoboTeam}, a simulated benchmark where agents control 4 to 16 robot arms in real time from camera images alone.
Its 36 tasks are long jobs that the arms must do together, such as passing timbers or building a picture from hundreds of cubes.
We start from the most flexible agent system, the adaptive system, where one agent decides when to fork delegates.
With it, \mbox{gpt-6.1-sol} and \mbox{opus-5.5} complete 28 and 27 tasks, and four other models at most 3.
But at a high price: a typical task takes \mbox{gpt-6.1-sol} 3.5 hours and 218 million input tokens, and a third of its steps fail at least once and must be retried.
We then compare it with three fixed systems: single-agent, hierarchical, and decentralized.
All four complete a similar number of tasks but differ in time and tokens.
The hierarchical and decentralized systems are faster but use up to 1.9 times the tokens, and the single agent uses fewer tokens but takes 1.6 times as long, so the adaptive system lies between them.
Even so, \textbf{no model or system is yet fast, cheap, and reliable at once}.
RoboTeam measures these limits where earlier benchmarks cannot: long multi-arm jobs, camera images only, and a world that never pauses.
\end{abstract}

\section{Introduction}
\label{sec:intro}

In the Spider-Man stories, Doctor Octopus controls four extra arms as easily as his own hands.
Real robots are far from this, but multimodal agents are getting closer: frontier models can increasingly control robots through tools and generated code~\cite{saycan,codeaspolicies,faea,capx,showharness,bitterlesson,robotworld,firstexam}.
Vision-language-action (VLA) policies~\cite{rt2,openvla,pi0} typically map camera images directly to robot actions.
Agents can do more: they reason about a task~\cite{react}, write tools~\cite{voyager}, recover from failure~\cite{reflect}, and communicate~\cite{coela}.

\begin{figure}[t]
  \centering
  \includegraphics[width=\linewidth]{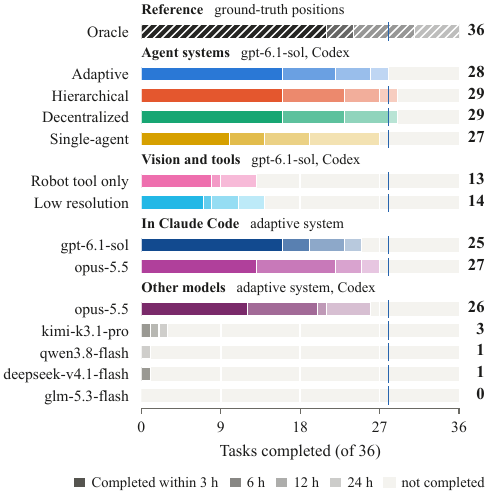}
  \caption{How many of the 36 tasks each condition completes. Darker means completed sooner. The blue line marks the adaptive system of \mbox{gpt-6.1-sol} in Codex.}
  \label{fig:leaderboard}
\end{figure}

Existing benchmarks capture these strengths only in part, and they fall short in three ways (\cref{tab:related}).
First, most test one or two arms on short tasks, whether the controller is a trained policy~\cite{rlbench,robotwin,bicoord} or a language-model agent~\cite{embodiedbench,robocoliseum,robotworld}.
Multi-robot studies compare agent systems but simplify the physical task~\cite{hmas,climbench,partnr}.
Second, many simulated benchmarks pause the world while an agent thinks~\cite{robotworld,orak}, which a real robot cannot do.
Third, many give agents the ground-truth positions of objects~\cite{roco,hmas,climbench,capx}, which a real robot must estimate from its sensors.
We therefore still do not know how far agents are from controlling many arms at once, or which agent system suits them.

To study this question, we present RoboTeam, a simulated benchmark where agents control 4 to 16 arms in real time from camera images alone (\cref{fig:overview}).
Its 36 tasks are long multi-arm tasks of four types modeled on real multi-robot work: sharing parts in an assembly cell, laying bricks, carrying long parts, and turning valves (\cref{sec:design}).
Each comes at three team sizes and three difficulty levels.
They do not demand fast reactions, because language models take seconds to respond~\cite{orak,bitterlesson}.
Agents move the arms through one robot tool with simple commands, and a sandbox keeps them away from the simulator and the evaluator, which prevents cheating~\cite{impossiblebench,evilgenie,metrhack}.

\begin{table}[t]
  \caption{RoboTeam compared with related benchmarks. The first seven test LLM or VLM agents and the next four test learned policies. Real-time: the world never pauses. Vision only: agents get no ground-truth object positions. Run length: typical time of a run, or its time limit where marked $\le$. \checkmark\ yes, $\sim$ partly, $\times$ no, $\approx$ our estimate, --- single query.}
  \label{tab:related}
  \centering
  \footnotesize
  \setlength{\tabcolsep}{3.4pt}
  \begin{tabular}{@{}lccccc@{}}
    \toprule
    & Max & Multi- & Real- & Vision & Run length \\
    Benchmark & arms & agent & time & only & \\
    \midrule
    RoboColiseum~\cite{robocoliseum} & 2 & $\times$ & $\times$ & \checkmark & 33 min \\
    RobotWorld~\cite{robotworld} & 2 & $\times$ & $\times$ & $\sim$ & 26 min \\
    EmbodiedBench~\cite{embodiedbench} & 1 & $\times$ & $\times$ & $\times$ & ${\approx}$1 min \\
    PARTNR~\cite{partnr} & 1 & \checkmark & $\times$ & $\sim$ & 36 min \\
    RoCoBench~\cite{roco} & 3 & $\sim$ & $\times$ & $\times$ & ${\approx}$10 min \\
    VIKI-Bench~\cite{vikir} & 2 & $\times$ & $\times$ & $\sim$ & --- \\
    Embodied First Exam~\cite{firstexam} & 2 & $\times$ & \checkmark & \checkmark & ${\le}$4 h \\
    \midrule
    RoboTwin~\cite{robotwin} & 2 & $\times$ & $\times$ & \checkmark & ${\approx}$2 min \\
    RoboFactory~\cite{robofactory} & 4 & \checkmark & $\times$ & \checkmark & ${<}$1 min \\
    VLABench~\cite{vlabench} & 1 & $\times$ & $\times$ & $\sim$ & ${\approx}$1 min \\
    ReflexBench~\cite{reflexbench} & 1 & $\times$ & $\sim$ & \checkmark & ${<}$0.1 min \\
    \midrule
    \textbf{RoboTeam (ours)} & \textbf{16} & \checkmark & \checkmark & \checkmark & \textbf{${\le}$24 h} \\
    \bottomrule
  \end{tabular}

\end{table}

We start from the most flexible agent system~\cite{mast,pappu,hmas,scalingagents}.
In this adaptive system, one controller decides for itself when to fork delegates and hand them arms, and it takes the arms back when their jobs end.
With this system we test six models, and we then compare it with three fixed systems: single-agent, hierarchical, and decentralized.
Further conditions change the harness, what the agent sees, or the tools it may use, which gives 14 conditions (\cref{fig:leaderboard}).

\textbf{The best models complete most tasks, but at a high price.}
With the adaptive system, \mbox{gpt-6.1-sol} and \mbox{opus-5.5} complete 28 and 27 of the 36 tasks, while the other four models complete at most 3.
A typical task takes \mbox{gpt-6.1-sol} 3.5 hours and 218 million input tokens.
An atomic step is the smallest unit of progress, such as placing one cube, and 34\% of these steps fail at least once and must be retried.

\textbf{Other agent systems trade time for tokens.}
The four systems complete a similar number of tasks.
The hierarchical and decentralized systems are faster than the adaptive system but use 1.9 and 1.7 times its tokens, and the single agent uses fewer tokens but takes 1.6 times as long.
The adaptive system lies between them in both time and tokens.

\textbf{What limits the agents is recovering positions from images.}
Given the ground-truth object positions instead of camera images, \mbox{gpt-6.1-sol} completes all 36 tasks.
From images alone it needs two things, and losing either one costs about half of the completed tasks.
With low-resolution images it completes 14 tasks, and without the code tools that it uses to measure positions in images, 13.

The two best models also work differently.
\mbox{gpt-6.1-sol} divides its work among delegates, and \mbox{opus-5.5} works mostly as one agent that computes more, in either harness.
No model or agent system is yet fast, cheap, and reliable at once, and RoboTeam measures all three.
We will release the benchmark, the runtime, and all run records.

\begin{figure}[t]
  \centering
  \includegraphics[width=\linewidth]{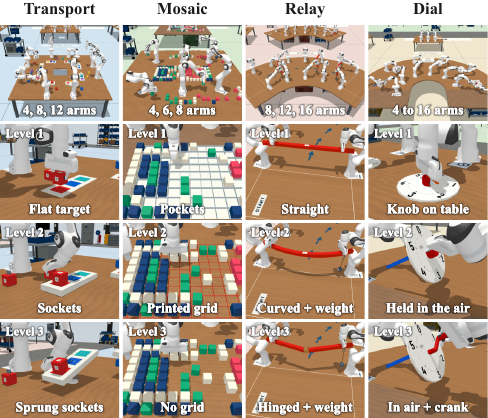}
  \caption{The four task types. Top row: the whole scene and its team sizes. Other rows: what changes from Level 1 to Level 3.}
  \label{fig:tasks}
\end{figure}

\section{Related work}
\label{sec:related}

\paragraph{Agents for robot control.}
Frontier agents now control robots through software agent frameworks~\cite{faea}, generated code~\cite{capx}, and simple action interfaces~\cite{showharness,bitterlesson}.
RobotWorld~\cite{robotworld}, RoboColiseum~\cite{robocoliseum}, and EmbodiedBench~\cite{embodiedbench} evaluate such agents broadly, but one model makes the decisions and controls at most two arms.
The concurrent Embodied First Exam~\cite{firstexam} also has a setting where the world keeps running, but one coding agent handles each task.

\paragraph{VLA and other policy benchmarks.}
VLA policies are usually trained from demonstrations~\cite{rt2,openvla,pi0} and react many times per second, so their benchmarks focus on skills of hundreds of control steps~\cite{robotwin,bicoord,vlabench}.
These benchmarks do not ask a model to divide work, talk to teammates, or write tools, so RoboTeam complements them.
ReflexBench~\cite{reflexbench} also keeps its world moving, but it tests whether a fast policy reacts in time.
RoboTeam tests the opposite case, whether slow, deliberate agents can still finish long tasks in a moving world.

\paragraph{Multi-agent and multi-robot systems.}
RoCo~\cite{roco} has one language model per robot discuss plans, with at most three arms and accurate state.
Chen \etal~\cite{hmas} compare centralized, decentralized, and hybrid teams, but their largest teams work in abstract 2D tasks.
CLiMBench~\cite{climbench} forms dynamic subgroups using coordinate state and magnetic assembly, and PARTNR~\cite{partnr} uses skills that hide contact.
RoboFactory~\cite{robofactory} and VIKI-Bench~\cite{vikir} study multi-robot cooperation with imitation policies and vision-language planners (\cref{tab:related}).
Software studies compare communication structures~\cite{multiagentbench} and create sub-agents on demand~\cite{aorchestra,pappu}.
RoboTeam tests these ideas with agents that see only through cameras and must share arms while the world keeps moving.

To our knowledge, RoboTeam is the first benchmark where multiple agents control many arms while the world keeps running as they think.

\begin{figure}[t]
  \centering
  \includegraphics[width=\linewidth]{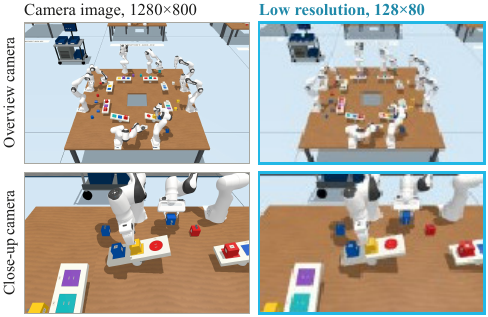}
  \caption{One Transport scene from two cameras at $1280\times800$ pixels and at the low resolution of one of our conditions (\cref{sec:protocol}), rendered with the simulator's OpenGL renderer.}
  \label{fig:vision}
\end{figure}

\section{The RoboTeam benchmark}
\label{sec:design}

\subsection{Tasks}

RoboTeam has 36 tasks: four task types, each at three team sizes and three difficulty levels (\cref{fig:tasks}).
Team size adds arms and work to divide.
Difficulty makes the same job physically or visually harder.

\smallskip\noindent\textbf{Transport.}
As in an assembly cell where robots share parts and must not collide~\cite{roboballet,multiarmtamp,wrede2024dual}, the arms deliver every item to its target board.
Items rest on flat targets (Level 1), plug into sockets (Level 2), or plug into sockets on a sprung board (Level 3).

\begin{figure*}[t]
  \centering
  \includegraphics[width=\linewidth]{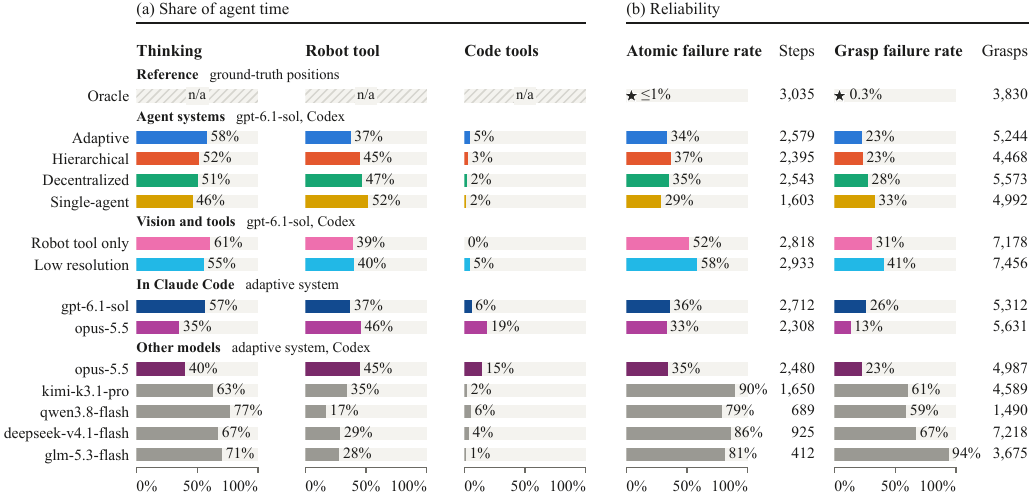}
  \caption{How every condition spends its time and how often it fails. (a) Share of agent time spent thinking, in the robot tool, and in code tools. (b) Median atomic failure rate and the share of grasps that fail (\cref{sec:metrics}), with total steps and grasps. The star marks the Oracle.}
  \label{fig:shares}
\end{figure*}

\smallskip\noindent\textbf{Mosaic.}
As in robotic bricklaying~\cite{brickwall,handover,dindorf2026bricklaying}, the arms build a reference picture from colored cubes, with one cube in each cell.
The cells are pockets (Level 1), printed lines (Level 2), or unmarked (Level 3).

\smallskip\noindent\textbf{Relay.}
As in a team that carries long parts and passes them along~\cite{longbeam,dogar,handover}, the arms move long timbers to a common goal.
Every handover between arms must happen in the air, and a dropped timber must be picked up again.
The timber is straight (Level 1), curved with a hanging weight (Level 2), or hinged in the middle with a hanging weight (Level 3).

\smallskip\noindent\textbf{Dial.}
As in turning valves~\cite{xing2024valve} that spring back unless held~\cite{oshadeadman}, the arms find a hidden code by turning spring-loaded knobs and holding them.
Their only guide is a display of the best number of correct dials so far.
At Level 1 the knobs sit on the table.
At Level 2 one arm must hold each module in the air while another turns its knob, and at Level 3 the knob becomes a crank.

\smallskip
Two principles guided these tasks.
First, teamwork must have value: arms depend on each other through shared space, shared objects, and mutual support, and we exclude tasks that each arm could finish alone.
This teamwork is between arms, and one agent may control all of them.
Second, tasks must be hard but feasible: every action is physically possible for the Franka Emika Panda arm and its parallel gripper~\cite{menagerie}.

\subsection{Environment}
\label{sec:environment}

Three choices define the environment: real time, what agents perceive, and how they act.

The first choice is real time: the world advances with the clock and never waits for an agent.
A real robot would not wait for a model, and pausing for one agent would also stop its teammates.
The world is simulated in MuJoCo~\cite{mujoco} with a 2\,ms physics step.

The second choice is what agents perceive.
Some benchmarks give agents ground-truth object positions~\cite{roco,hmas,climbench} or offer them as a separate setting~\cite{capx}, and agents do well with this access~\cite{faea}.
A real robot must estimate positions from its sensors, so our agents get only two kinds of live data.
The first is RGB images from fixed cameras, which agents request by name and which each camera renders at $1280\times800$ pixels (\cref{fig:vision}).
MuJoCo's OpenGL renderer draws an image in about 30\,ms on a laptop GPU, in a separate process, so looking does not slow the world.
A task has 14 to 69 cameras: two at every arm, four overviews of the scene, and close-ups of the task objects.
The task file also gives the position and the field of view of every camera, so agents can relate several views.
The second is the state of the arms, such as the pose and width of each gripper.
Agents never receive object positions, depth, or contact information.

\begin{figure}[t]
  \centering
  \includegraphics[width=\linewidth]{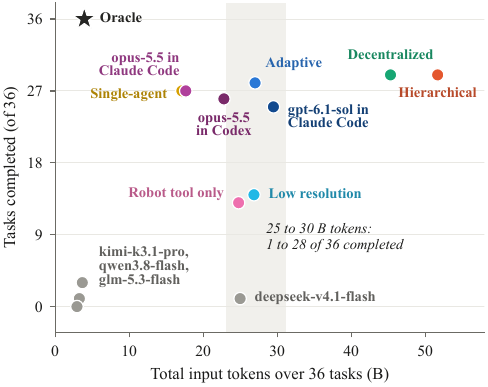}
  \caption{Tasks completed against input tokens used, one point per condition. The five conditions in the shaded band used about 25 to 30 billion (B) tokens each but completed 1 to 28 of the 36 tasks.}
  \label{fig:cost}
\end{figure}

The third choice is how agents act.
Earlier interfaces have two problems.
When agent code drives the simulator directly~\cite{faea,capx}, it can also read hidden state.
When agents call fixed high-level skills~\cite{partnr,capx}, the skills hide low-level failures.
We instead give every model one robot tool, and arms move only through it.
Its commands are simple: move an arm to a pose, or open, close, or grasp with a gripper.
This keeps agents away from the simulator and makes every failed motion or grasp visible.
Unless a condition removes them, agents also have code tools: a shell, file editing, and their own Python code (\cref{fig:overview}).
They use these tools, for example, to measure object positions in an image.
A sandbox keeps agents away from the evaluator, hidden task data, and our code.
It also blocks simulators, motion planners, and deep learning libraries, so agents cannot rebuild the world offline or bring in trained models.

\subsection{Agent systems}
\label{sec:systems}

We compare four agent systems (\cref{fig:overview}).
The adaptive system is the most flexible, because the agents decide during a task how to divide the work.
It starts as one controller that holds every arm.
When work can be shared, the controller forks itself into delegates and hands them arms.
When a delegate finishes its job, it returns its arms and ends.
Two rules make this work in real time.
A delegate is a fork of the controller. Unlike sub-agents with curated or separate context~\cite{aorchestra,anthropicmas}, it starts with the controller's full history and does not have to relearn the task while the arms wait.
Each arm belongs to one agent at a time, so two agents never command the same arm.
The other three systems keep a fixed structure during a task.
In the single-agent system, one agent controls every arm.
Following earlier work~\cite{hmas,scalingagents}, the hierarchical system has one worker per arm under a controller.
The decentralized system has the same workers without a controller, so they coordinate directly.

In every system, agents can also write lessons to one shared skill file~\cite{voyager,reflexion,expel}, which starts empty in every run.

\subsection{Metrics}
\label{sec:metrics}

RoboTeam does not end a task at its first failed step, as scoring a single attempt would.
Agents may retry, so a task that ends correctly can still contain many failures.
We therefore report whether a task was completed, how often its steps and grasps failed on the way, and what the run cost in time and input tokens.

\textbf{Completion.}
A task is completed when its final state is correct, for example when every Mosaic cube rests on its cell.
We also count the tasks completed within 3, 6, 12, and 24 hours, which shows how fast a condition works.
A 3-hour budget is a cheaper measure, since a task not done by then is already a weak result.

\textbf{Atomic failure rate.}
An atomic step is the smallest unit of progress.
In Mosaic it is one cube placed on its cell.
In Transport it is one item delivered to its target or handed to another arm.
In Relay it is one arm taking hold of a timber in a handover.
In Dial it is one knob turned to a new position.
A step fails if any attempt at it goes wrong, for example when an object is dropped.
We find these failures after a run in the records of the robot tool and the simulator, not in the agents' reports.
A step still counts as failed even if it is completed later.
The atomic failure rate of a run is the share of its steps that fail at least once, and we report its median over the 36 tasks.

\textbf{Grasp failure rate.}
A grasp fails when the gripper closes on a width that does not match the object.
The grasp failure rate of a condition is the share of all its grasps that fail.

\section{Experiments}
\label{sec:results}

\subsection{Setup}
\label{sec:protocol}

We test six models: \mbox{gpt-6.1-sol}, \mbox{opus-5.5}, \mbox{kimi-k3.1-pro}, \mbox{qwen3.8-flash}, \mbox{deepseek-v4.1-flash}, and \mbox{glm-5.3-flash}.
All use high reasoning effort and a 300{,}000-token context window.
Agents run in a coding-agent harness, either Codex~\cite{codex} or Claude Code~\cite{claudecode}.
Our runtime connects the harness to the world and records every call.
In both harnesses, agents get the same robot tool, instructions, camera images, and sandbox.

Unless a condition changes them, agents run \mbox{gpt-6.1-sol} in Codex with code tools.
In all, there are 14 conditions in four groups and a reference (\cref{fig:leaderboard}), and each runs the 36 tasks once.
The first group runs the four agent systems of \cref{sec:systems}, and all other conditions except the Oracle use the most flexible of them, the adaptive system.
The second changes what the agent perceives or may use: \emph{robot tool only} removes the code tools, and \emph{low resolution} reduces every camera image to $128\times80$ pixels (\cref{fig:vision}).
\mbox{gpt-6.1-sol} and \mbox{opus-5.5} run in their providers' harnesses, Codex and Claude Code, and the other four models in Codex.
To test the harness, we swap the two: the third group runs \mbox{gpt-6.1-sol} and \mbox{opus-5.5} in Claude Code, and the fourth runs the other models, \mbox{opus-5.5} among them, in Codex.
The \emph{Oracle} uses the single-agent system and receives the ground-truth position of every object instead of camera images (\cref{tab:allmetrics}).
It is a reference for what the tasks allow, and its world still runs in real time.
We also ran the adaptive system twice more on the 12 Level 3 tasks.

Agents are told to keep trying and are not told their time budget.
A run ends when the agents declare the task done or when the 24-hour budget runs out.
The budget counts only active time: when the infrastructure pauses a run, for example during a network outage, the world, the clock, and the agents stop together.
Agents may also stop a run themselves, but only after long and repeated failure at one step.

\subsection{Main results}
\label{sec:overview}

\textbf{The best models complete most tasks, and the other four almost none.}
\Cref{fig:leaderboard} shows how many tasks each of the 14 conditions completes.
\mbox{gpt-6.1-sol} completes 27 to 29 of the 36 tasks under the four agent systems, and \mbox{opus-5.5} completes 27 in Claude Code and 26 in Codex.
The other four models complete at most 3 tasks and no Level 3 task.
The Oracle completes all 36, so every task is feasible.

This gap between models is not a matter of budget (\cref{fig:cost}).
\mbox{deepseek-v4.1-flash} uses 25.0 billion input tokens and 316 hours.
This is close to the 27.0 billion tokens and 258 hours of \mbox{gpt-6.1-sol}, but it completes one task.

\begin{figure}[t]
  \centering
  \includegraphics[width=\linewidth]{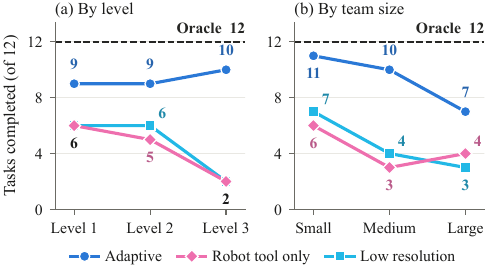}
  \caption{Tasks completed by \mbox{gpt-6.1-sol} in Codex at each level and each team size (12 tasks each). The Oracle gets ground-truth object positions and completes all 12.}
  \label{fig:levels}
\end{figure}

For the adaptive system, team size makes tasks harder than level does (\cref{fig:levels}).
By level, it completes 9, 9, and 10 of the 12 tasks.
By team size, it completes 11, 10, and 7 on small, medium, and large teams.
The levels matter much more once the agent sees less or computes less (\cref{sec:vision}).

\textbf{Repeated runs.}
Two more runs of the adaptive system on the 12 Level 3 tasks complete 8 and 9 of them, against 10 in the first run, and no task fails in all three (\cref{sec:supp-failures}).
A second run can thus differ by one or two tasks, so we read only larger differences as real.

\subsection{The price of completion}
\label{sec:tools}

\begin{figure}[t]
  \centering
  \includegraphics[width=\linewidth]{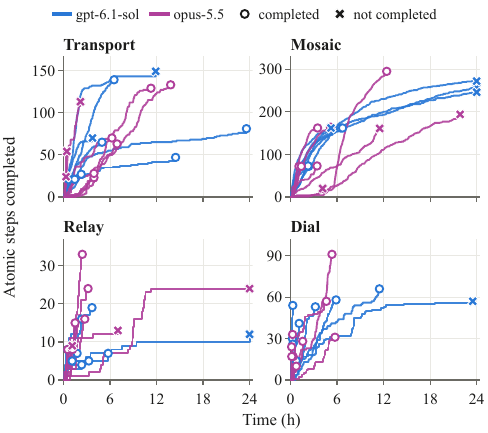}
  \caption{Progress of every task over time for \mbox{gpt-6.1-sol} in Codex and \mbox{opus-5.5} in Claude Code, by task type. Each line is one run and rises when an atomic step is completed. A circle marks a completed task and a cross one that was not completed. Each task type has its own scale.}
  \label{fig:steps}
\end{figure}

\textbf{Tasks are completed only by retrying, and they take hours and many tokens.}
\Cref{fig:steps} shows every run of \mbox{gpt-6.1-sol} in Codex and \mbox{opus-5.5} in Claude Code.
For \mbox{gpt-6.1-sol}, the median atomic failure rate is 13\% in Dial, 23\% in Mosaic, 37\% in Transport, and 100\% in Relay.
In the median Relay task, every handover fails at least once.
The task has only 7 steps, however, and 8 of the 9 Relay tasks are completed.
In Mosaic most steps work, but the median task has 162 of them, and only 5 of the 9 tasks are completed.
In simulation a dropped cube can simply be picked up again, but on real robots a failed step can break a part or stop a line.
The atomic failure rate therefore says more about practical use than the number of completed tasks (\cref{sec:supp-failures} analyzes the eight failed tasks).

Time is the second cost.
A median task takes \mbox{gpt-6.1-sol} 3.5 hours and 218 million input tokens, although each step is as simple as moving one cube.
Its agents spend 58\% of their time thinking, 37\% in the robot tool, and 5\% in code tools (\cref{fig:shares}a).
In one run's timeline, at least one agent is thinking during 90\% of the time, but at least one arm moves during only 43\% (\cref{fig:overview}, bottom right).

In every condition, at least 90\% of input tokens are cached context, so agents spend much of their budget rereading what they already know.
Since arms sit idle most of the time, faster decisions would shorten tasks more than faster motion would.

\begin{figure}[t]
  \centering
  \includegraphics[width=\linewidth]{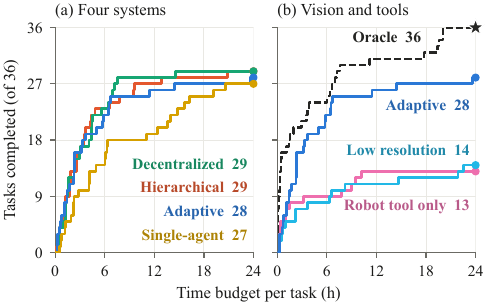}
  \caption{Tasks completed by \mbox{gpt-6.1-sol} in Codex as the time allowed per task grows. (a) The four agent systems. (b) The adaptive system against the Oracle, which gets ground-truth object positions, and against the two conditions that take away code tools or image resolution.}
  \label{fig:time}
\end{figure}

\subsection{Agent systems}
\label{sec:systemresults}

\textbf{Other agent systems trade time for tokens, and the adaptive system lies between them.}
With \mbox{gpt-6.1-sol}, the four systems complete 27 to 29 tasks, and 20 tasks are completed by all four.
The clear difference is speed (\cref{fig:time}a).
Within 6 hours, the three multi-agent systems complete 22 or 23 tasks and the single agent 14.
On the tasks that two systems both complete, the three multi-agent systems are 2.3 to 2.6 times as fast as the single agent in the median task.
The gap grows with team size: on the largest teams, the median task takes the single agent 22.0 hours and the hierarchical system 5.5 hours.

\begin{figure}[t]
  \centering
  \includegraphics[width=\linewidth]{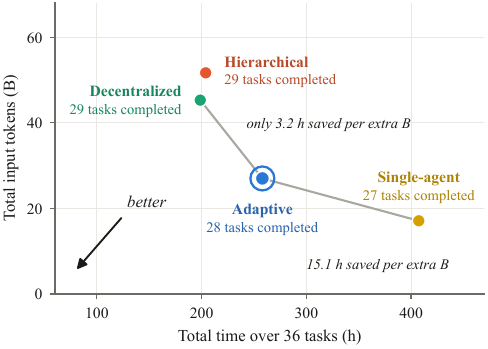}
  \caption{More agents save time but cost tokens. Each point is one agent system of \mbox{gpt-6.1-sol} in Codex, with its total time and input tokens in billions (B). The ring marks the adaptive system.}
  \label{fig:timetokens}
\end{figure}

This speed costs tokens, because more agents read more (\cref{fig:timetokens}).
The adaptive system sits between the two extremes: the hierarchical and decentralized systems use 1.9 and 1.7 times its tokens, and the single agent takes 1.6 times as long in total.
On the tasks that both complete, the hierarchical and decentralized systems are 1.3 and 1.2 times as fast as the adaptive system in the median task.
Total times include failed runs: 120 of the 258 hours of the adaptive system come from five tasks that reach the 24-hour limit.
The atomic failure rate is similar across the four systems (\cref{fig:shares}b).

Communication explains the cost (\cref{fig:overview}): the median task has 471 and 421 messages in the hierarchical and decentralized systems, and 184 in the adaptive system, which creates delegates only when work can be shared.

\begin{figure*}[t]
  \centering
  \includegraphics[width=\linewidth]{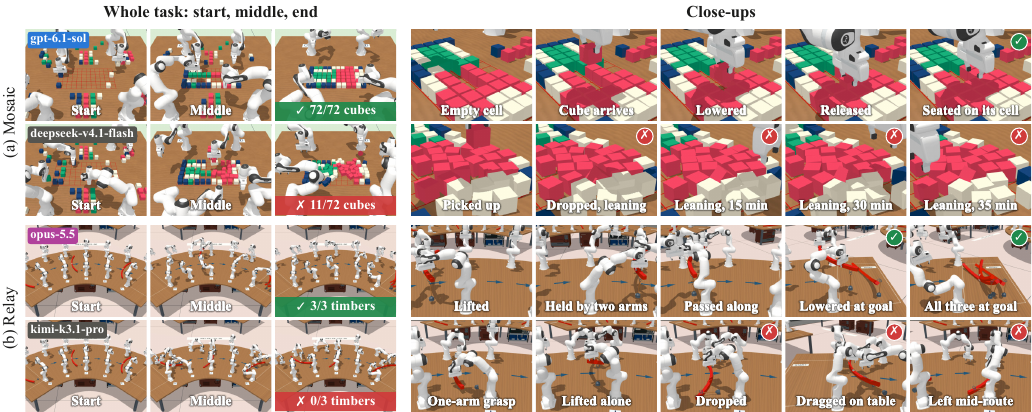}
  \caption{Two tasks as recorded, each with a model that completes it and one that does not. Left: the scene at the start, middle, and end. Right: close-ups of key moments, where a tick marks a correct state and a cross a faulty one.}
  \label{fig:qual}
\end{figure*}

\subsection{Vision and tools}
\label{sec:vision}

\textbf{The bottleneck is recovering positions from images.}
With ground-truth positions every task is completed, and without them agents need both sharp images and code tools.
Three conditions show this by changing what \mbox{gpt-6.1-sol} perceives or may use (\cref{fig:time}b).

The Oracle completes all 36 tasks, and only 0.3\% of its grasps fail (\cref{fig:shares}b).
It takes a median of 1.45 hours per task.
With the same model and code tools but only camera images, the adaptive system must find the objects itself.
It completes 28 tasks, and 23\% of its grasps fail.
This gap suggests that the agent is limited by what it can see more than by planning or control.

Without ground-truth positions, taking away code tools cuts the completed tasks from 28 to 13, and taking away sharp images cuts them to 14.
Across the five conditions that keep both, \mbox{gpt-6.1-sol} completes 25 to 29 tasks.
In paired tests over the 36 tasks (\cref{sec:supp-results}), both drops are significant ($p<0.005$ after correction), and no two of those five conditions differ detectably.
With the robot tool only, the agent sees the same images but loses computation, scripted control, and its working files, so this condition shows what code tools contribute as a whole.
With low resolution, it keeps its code tools but sees every image at $128\times80$ pixels (\cref{fig:vision}).
The two losses look alike: both conditions complete no Mosaic task, and both fall as the level rises.
At Level 3 each completes 2 of the 12 tasks, against 10 with both (\cref{fig:levels}a).
Their grasps also fail more often, 31\% and 41\% of the time against 23\% (\cref{fig:shares}b).

Both are needed for one reason.
An agent can act on an object only if it knows the object's position to within millimeters, and it gets this position by measuring images.
In 31 of its 36 adaptive runs, \mbox{gpt-6.1-sol} writes code for this.
One script, for example, finds an object by its color in two views and triangulates its 3D position.
Low resolution removes the detail that such measurements need.
The notes that agents write during these runs say what is lost (\cref{sec:supp-lowres}).
One pixel covers about 14.5\,mm, more than twice the 6\,mm tolerance for placing a cube.
Neighboring cubes of one color merge into a single patch, and the digits on a dial become hard to tell apart.
These notes point to measurement, not recognition, as the bottleneck: the hard part is turning pixels into positions accurate enough to act on.

\subsection{Models and harness}
\label{sec:modelresults}

\textbf{The two best models work in different ways, and each keeps its way of working in both harnesses.}
With the adaptive system, \mbox{gpt-6.1-sol} completes 28 tasks in Codex and \mbox{opus-5.5} completes 27 in Claude Code, and each completes some that the other does not.
\mbox{gpt-6.1-sol} divides the work, and its controller delegates 13 times in the median task.
\mbox{opus-5.5} rarely delegates, and in 20 of the 36 tasks it never does.
Instead, it thinks less and computes more.
It spends 35\% of agent time thinking and 19\% in code tools, against 58\% and 5\% for \mbox{gpt-6.1-sol} (\cref{fig:shares}a).
Working mostly as one agent, it needs 176 hours for the 36 tasks, less than half the time of the single agent of \mbox{gpt-6.1-sol}.
The median atomic failure rates of the two models are almost the same, at 34\% and 33\% (\cref{fig:shares}b).

Both models keep their way of working in the other harness.
The median task of \mbox{gpt-6.1-sol} has 184 messages in Codex and 115 in Claude Code, and that of \mbox{opus-5.5} has none in either.
\mbox{opus-5.5} spends 35\% and 40\% of agent time thinking, against 58\% and 57\% for \mbox{gpt-6.1-sol} (\cref{fig:shares}a).
In Codex, \mbox{opus-5.5} drives all 12 arms of a Transport task itself, noting that neighboring arms share handover positions.
The four combinations complete 25 to 28 tasks, and of the reliability measures only the grasp failure rate follows the harness: \mbox{opus-5.5} fails 13\% of its grasps in Claude Code and 23\% in Codex (\cref{fig:shares}b).

The other four models fail at basic manipulation.
Their median atomic failure rate is 79--90\%, and 59--94\% of their grasps fail (\cref{fig:shares}b).
They spend the largest share of their time thinking, 63--77\%.
Repeated failure rarely changes their behavior, however, and one \mbox{deepseek-v4.1-flash} arm failed 51 grasps in a row.
\Cref{fig:qual} gives two examples.
In Mosaic, \mbox{deepseek-v4.1-flash} leaves a dropped cube leaning on its neighbors, and in Relay, \mbox{kimi-k3.1-pro} lifts a timber with one arm and drops it.

\section{Conclusion}
\label{sec:conclusion}

In simulation, RoboTeam shows that the best agents can complete hours-long work across many robot arms, but only at a high price.
With the adaptive system, \mbox{gpt-6.1-sol} and \mbox{opus-5.5} complete 28 and 27 of the 36 tasks, while the other four models still fail at basic manipulation.
Even the best model needs hours and hundreds of millions of tokens per task, and it retries a third of its steps.
Fixed agent systems do not remove this price: they trade time for tokens with no detectable difference in completion, and the adaptive system lies between them.
What holds the agents back is recovering accurate positions from images: with ground-truth positions every task is completed, and without sharp images or code tools only about half as many are.
Our cross-harness runs suggest that how a model divides its work depends on the model more than on the harness.
No model or agent system is yet fast, cheap, and reliable at once, and making them so is the next challenge.

\textbf{Limitations.}
Our results come from simulation, and transfer to real robots is future work.
The 36 tasks are variants of only four task types.
Because one condition takes billions of tokens, each ran once apart from the repeats of \cref{sec:overview}, so differences such as 28 versus 29 tasks are not reliable.
We tested no further model families, no depth images or learned perception, only one reduced resolution, and no paused-world control.

{
    \small
    \bibliographystyle{ieeenat_fullname}
    \bibliography{main}
}

\setcounter{topnumber}{99}
\clearpage
\setcounter{section}{0}\renewcommand{\thesection}{\Alph{section}}\renewcommand{\theHsection}{supp.\Alph{section}}
\setcounter{figure}{0}\renewcommand{\thefigure}{S\arabic{figure}}\renewcommand{\theHfigure}{supp.\arabic{figure}}
\setcounter{table}{0}\renewcommand{\thetable}{S\arabic{table}}\renewcommand{\theHtable}{supp.\arabic{table}}
\definecolor{skillbg}{gray}{0.955}\definecolor{skillrule}{gray}{0.70}
\newsavebox{\suppcodebox}%
\providecommand{\skillbox}[1]{{\setlength{\fboxsep}{5pt}\noindent\fcolorbox{skillrule}{skillbg}{\parbox{\dimexpr\linewidth-2\fboxsep-2\fboxrule\relax}{\footnotesize\sffamily\raggedright #1}}}}
\newlength{\suppmsgh}%
\providecommand{\skillboxh}[2]{{\setlength{\fboxsep}{5pt}\noindent\fcolorbox{skillrule}{skillbg}{\parbox[t][#1][t]{\dimexpr\linewidth-2\fboxsep-2\fboxrule\relax}{\footnotesize\sffamily\raggedright #2}}}}
\setcounter{topnumber}{3}\setcounter{dbltopnumber}{3}\renewcommand{\topfraction}{0.9}\renewcommand{\dbltopfraction}{0.9}\renewcommand{\textfraction}{0.08}
\setlength{\textfloatsep}{15pt plus 2pt minus 3pt}\setlength{\dbltextfloatsep}{15pt plus 2pt minus 3pt}
\newpage
\twocolumn[{%
  \centering
  {\Large\textbf{Appendix}\\
  \vspace{1.0em}}
  \captionsetup{type=figure}
  \includegraphics[width=\linewidth]{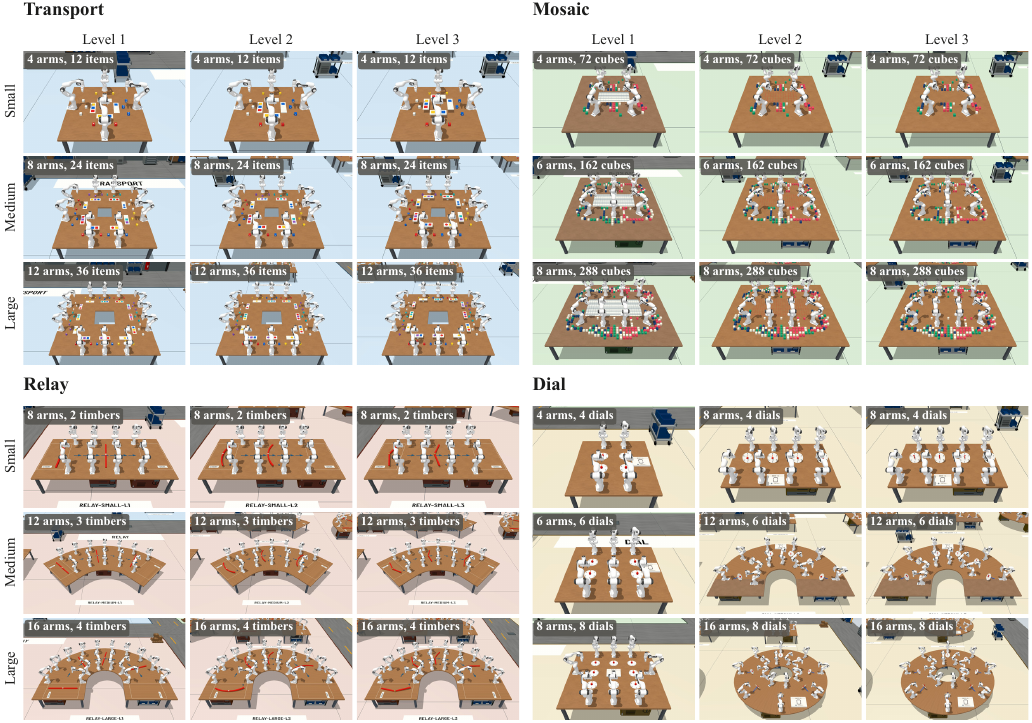}
  \caption{The 36 tasks at their start. Each block is one task type. Rows are the team sizes and columns are the levels. The label on a picture gives the number of arms and the amount of work.}
  \label{fig:tasksheet}
  \vspace{1.2em}
}]

\begin{figure*}[t]
  \centering
  \includegraphics[width=\linewidth]{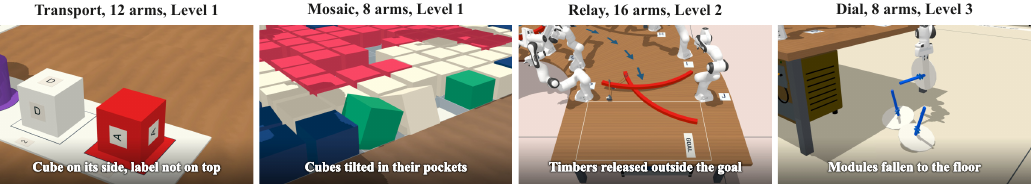}
  \caption{Four of the eight tasks that \mbox{gpt-6.1-sol} with the adaptive system did not complete, shown at the end of the run. In Transport, a cube lies on its side on its target. Its labeled top face points sideways and a blank face is up, unlike on the cube beside it. In Mosaic, two of the four tilted cubes sit crooked in their pockets. In Relay, two released timbers lie mostly outside the white goal rectangle. In Dial, two modules have fallen from the table to the floor, and an arm reaches down with a third.}
  \label{fig:failures}
\end{figure*}

\begin{table*}[t]
  \caption{The eight tasks that \mbox{gpt-6.1-sol} with the adaptive system did not complete. The upper three ended when the agents declared the task done. The lower five ended at the 24-hour limit.}
  \label{tab:failures}
  \centering
  \footnotesize
  \setlength{\tabcolsep}{5pt}
  \begin{tabular*}{\linewidth}{@{\extracolsep{\fill}}llll@{}}
    \toprule
    Task & At the end & End of the run & What is wrong \\
    \midrule
    Transport, 8 arms, Level 1 & 23 of 24 items & declared done, 3.7 h & One cube is 5.3\,mm off center and 4\,mm outside its target square \\
    Transport, 12 arms, Level 1 & 30 of 36 items & declared done, 11.9 h & Two cubes lie on a side face, and four cylinders are 5.5 to 8.1\,mm off center \\
    Mosaic, 6 arms, Level 2 & 160 of 162 cubes & declared done, 5.2 h & Two cubes are 10 and 11\,mm off center, against a limit of 6\,mm \\
    \midrule
    Mosaic, 8 arms, Level 1 & 256 of 288 cubes & 24 h limit & 26 cells empty, 4 cubes tilted in their pockets, 1 of the wrong color, 1 still held \\
    Mosaic, 8 arms, Level 2 & 241 of 288 cubes & 24 h limit & 40 cells empty, 7 cubes off center or rotated \\
    Mosaic, 8 arms, Level 3 & 272 of 288 cubes & 24 h limit & 11 cells empty, 5 cubes 6.1 to 7.1\,mm off center \\
    Relay, 16 arms, Level 2 & 0 of 4 timbers & 24 h limit & Two timbers released short of the goal, two still carried \\
    Dial, 8 arms, Level 3 & 1 of 4 dials at best & 24 h limit & Two of the four modules lie on the floor \\
    \bottomrule
  \end{tabular*}

\end{table*}

\section{The 36 tasks}
\label{sec:supp-tasks}

\Cref{fig:tasksheet} shows each of the 36 tasks at its start.
The label on a picture gives the number of arms and the amount of work, such as the number of cubes.
In Dial, Level 1 uses one arm per dial.
Levels 2 and 3 use two arms per dial, because one arm must hold the module while another turns its knob.

The task file tells the agents when a task counts as completed.
The rules differ by task type.

\smallskip\noindent\textbf{Transport.}
Every item must stand upright in its target.
An item carries its label on its top face and on two opposite side faces, so that cameras above and beside it can read it.
The top face must face up, with a tilt of at most 5 degrees.
The item must sit at the right height, within 1.5\,mm, and it must be released and at rest.
At Level 1, the whole item must also lie inside the 70\,mm square of its target.

\smallskip\noindent\textbf{Mosaic.}
At Level 1, every cube must sit flat inside its pocket.
At Levels 2 and 3, the center of every cube must lie within 6\,mm of the center of its cell.
The cubes must be at rest.

\smallskip\noindent\textbf{Relay.}
Every timber must lie in the goal rectangle, and every arm must have released it.
On the way, a timber must stay clear of the table and the floor, so every handover between arms happens in the air.
When a timber is dropped, the arms must pick it up again.

\smallskip\noindent\textbf{Dial.}
All dials must be within 4 degrees of their correct positions at the same time.

\section{Tasks the adaptive system did not complete, and repeated runs}
\label{sec:supp-failures}

With the adaptive system, \mbox{gpt-6.1-sol} completes 28 of the 36 tasks.
\Cref{tab:failures} lists the other eight, and \cref{fig:failures} shows four of them at the end of the run.
They fall into two groups.

\textbf{Declared done too early.}
In three tasks, the agents declare the task done while a few objects are outside the tolerance.
In the Transport task with 8 arms and in the Mosaic task with 6 arms, the agents had measured the faulty objects in the images and judged them to be in place.
They were not: one cube lies 4\,mm outside its target square, and two cubes lie 10 and 11\,mm from the centers of their cells.
In the Transport task with 12 arms, an arm knocked two cubes onto a side face early in the run, so their top faces no longer face up.
Later checks fitted the outline of each cube in the image, and the outline of a cube does not show which face is up.
Four cylinders were already off center when the arms placed them, and the later checks did not catch it.

\begin{table*}[t]
  \caption{The seven failures of the two repeated runs of \mbox{gpt-6.1-sol} with the adaptive system at Level 3. Run 2 completes 8 and run 3 completes 9 of the 12 tasks. Relay with 16 arms fails in both.}
  \label{tab:repeatfailures}
  \centering
  \footnotesize
  \setlength{\tabcolsep}{5pt}
  \begin{tabular*}{\linewidth}{@{\extracolsep{\fill}}llll>{\raggedright\arraybackslash}p{0.43\linewidth}@{}}
    \toprule
    Run & Task & At the end & End of the run & What is wrong \\
    \midrule
    2 & Transport, 12 arms & 5 of 36 items & stopped by agents, 0.9 h & A cube slipped from a gripper to the floor, out of reach. Three targets need this kind of cube, and there are only three \\
    2 & Mosaic, 8 arms & 259 of 288 cubes & 24 h limit & 177 cubes placed in the first 2 hours, then 4 to 8 per hour. 14 cells empty, 15 cubes up to 9\,mm off center \\
    2 & Relay, 16 arms & 2 of 4 timbers & declared done & Two timbers lie outside the goal rectangle \\
    2 & Dial, 8 arms & 0 of 4 dials & stopped by agents, 15 h & A module fell from the table at 0.4 h. Using the other modules as hooks to reach it, the agents dropped all four and never tried a code \\
    \midrule
    3 & Mosaic, 6 arms & 159 of 162 cubes & declared done & Three cubes are 9.2 to 9.7\,mm off center \\
    3 & Relay, 16 arms & 3 of 4 timbers & stopped by agents, 6.8 h & While correcting the position of one timber, an arm pushed it off the table \\
    3 & Dial, 16 arms & 2 of 8 dials shown & stopped by agents, 11 h & An arm moved a module unsteadily and flung it off the table \\
    \bottomrule
  \end{tabular*}

\end{table*}

\begin{table*}[t]
  \caption{All numbers of the 14 conditions over the 36 tasks. Time per task, messages, and tool calls are medians, and the parentheses give the middle half of the tasks (P25 to P75). Input and output tokens are totals, in billions (B) and millions (M). Cached is the share of input tokens that are read from the cache. The atomic failure rate is the median over the 36 tasks, again with the middle half. The last column gives the failed grasps, all grasps, and the share that failed. For the Oracle, time and tokens are counted as in every other condition, from when the agent receives the task until the run ends.}
  \label{tab:allmetrics}
  \centering
  \fontsize{7.4pt}{9pt}\selectfont
  \setlength{\tabcolsep}{1pt}
  \begin{tabular*}{\linewidth}{@{\extracolsep{\fill}}lrrrrrrrrrr@{}}
    \toprule
     & Tasks & Time per task & Total & Input & Output & & Messages & Tool calls & Atomic failure & Failed grasps \\
     & completed & (h) & time (h) & (B) & (M) & Cached & per task & per task & rate & of all grasps \\
    \midrule
    \multicolumn{11}{@{}l@{}}{\textbf{Reference} (\mbox{gpt-6.1-sol}, Codex, single agent, ground-truth positions)} \\
    Oracle & 36 & 1.45 (0.25--6.70) & 172.2 & 3.94 & 16.5 & 98.6\% & 0 (0--0) & 924 (373--1{,}919) & $\le$1\% & 13/3{,}830 (0.3\%) \\
    \midrule
    \multicolumn{11}{@{}l@{}}{\textbf{Agent systems} (\mbox{gpt-6.1-sol}, Codex)} \\
    Adaptive & 28 & 3.48 (1.44--7.90) & 258.0 & 27.01 & 65.7 & 97.9\% & 184 (59--706) & 1{,}576 (808--6{,}612) & 34\% (18--62\%) & 1{,}205/5{,}244 (23.0\%) \\
    Hierarchical & 29 & 3.09 (1.25--7.41) & 203.9 & 51.68 & 153.2 & 97.6\% & 471 (102--1{,}803) & 3{,}746 (1{,}428--12{,}996) & 37\% (19--58\%) & 1{,}047/4{,}468 (23.4\%) \\
    Decentralized & 29 & 4.20 (1.50--6.81) & 198.7 & 45.31 & 138.1 & 97.6\% & 421 (89--2{,}132) & 3{,}468 (1{,}253--15{,}478) & 35\% (17--56\%) & 1{,}545/5{,}573 (27.7\%) \\
    Single-agent & 27 & 8.70 (2.70--21.02) & 407.3 & 17.12 & 25.1 & 97.1\% & 0 (0--0) & 803 (527--4{,}025) & 29\% (14--49\%) & 1{,}666/4{,}992 (33.4\%) \\
    \midrule
    \multicolumn{11}{@{}l@{}}{\textbf{Vision and tools} (\mbox{gpt-6.1-sol}, Codex)} \\
    Robot tool only & 13 & 4.99 (1.17--10.59) & 286.0 & 24.79 & 49.6 & 96.4\% & 298 (89--813) & 2{,}786 (862--6{,}798) & 52\% (30--77\%) & 2{,}232/7{,}178 (31.1\%) \\
    Low resolution & 14 & 6.27 (2.23--13.17) & 308.0 & 26.86 & 72.3 & 97.1\% & 212 (69--883) & 1{,}703 (895--7{,}275) & 58\% (35--85\%) & 3{,}021/7{,}456 (40.5\%) \\
    \midrule
    \multicolumn{11}{@{}l@{}}{\textbf{In Claude Code} (adaptive system)} \\
    \mbox{gpt-6.1-sol} & 25 & 3.24 (1.73--9.02) & 267.9 & 29.49 & 97.1 & 90.1\% & 115 (31--451) & 1{,}690 (731--7{,}019) & 36\% (22--65\%) & 1{,}403/5{,}312 (26.4\%) \\
    \mbox{opus-5.5} & 27 & 3.29 (1.12--5.86) & 176.1 & 17.65 & 39.6 & 97.5\% & 0 (0--8) & 963 (430--3{,}864) & 33\% (22--50\%) & 747/5{,}631 (13.3\%) \\
    \midrule
    \multicolumn{11}{@{}l@{}}{\textbf{Other models} (adaptive system, Codex)} \\
    \mbox{opus-5.5} & 26 & 4.14 (1.62--7.22) & 223.7 & 22.80 & 45.6 & 95.2\% & 0 (0--10) & 1{,}069 (542--4{,}035) & 35\% (20--54\%) & 1{,}145/4{,}987 (23.0\%) \\
    \mbox{kimi-k3.1-pro} & 3 & 1.21 (0.65--3.29) & 82.4 & 3.67 & 10.5 & 95.8\% & 14 (8--34) & 848 (640--1{,}221) & 90\% (73--100\%) & 2{,}783/4{,}589 (60.6\%) \\
    \mbox{qwen3.8-flash} & 1 & 0.99 (0.45--3.37) & 75.1 & 3.27 & 100.9 & 94.8\% & 3 (0--22) & 790 (205--1{,}452) & 79\% (66--100\%) & 882/1{,}490 (59.2\%) \\
    \mbox{deepseek-v4.1-flash} & 1 & 3.01 (2.01--17.05) & 316.2 & 24.99 & 143.4 & 97.0\% & 12 (0--49) & 2{,}097 (1{,}196--9{,}931) & 86\% (78--99\%) & 4{,}860/7{,}218 (67.3\%) \\
    \mbox{glm-5.3-flash} & 0 & 1.03 (0.71--4.04) & 84.2 & 2.95 & 13.6 & 96.1\% & 3 (0--21) & 1{,}186 (910--1{,}774) & 81\% (76--97\%) & 3{,}453/3{,}675 (94.0\%) \\
    \bottomrule
  \end{tabular*}

\end{table*}

\textbf{Out of time.}
In five tasks, the 24 hours run out.
The three largest Mosaic tasks end with 241 to 272 of the 288 cubes in place.
Their work slows as the picture fills.
In the first six hours the arms place 154 to 168 cubes, and in the last six hours they place 14 to 25.
By then the agents exchange 36 to 71 messages for every cube placed, against 5 or 6 at the start.
Most of the time passes between motions, while the agents check planned motions and wait for each other.

In the Relay task, the agents release two timbers after computing that each lies in the goal rectangle.
The goal rectangle is rotated by 90 degrees on the table, and their computation left this rotation out.
Neither timber lies in the goal rectangle.
The other two timbers are still in the air.

In the Dial task, a module slips out of a grasp after about five hours and falls from the table to the floor.
The code cannot be entered without it.
The agents spend the rest of the run trying to lift it back with another module as a tool, and a second module falls.
Before the fall, they made two valid trials of the code.

Completion is strict, so a task with one object out of place counts as not completed.
Counted by objects, the six Transport and Mosaic tasks among the eight end with 982 of their 1{,}086 objects in place, which is 90\%.

\textbf{What four of the eight tasks share.}
In four of the eight tasks, the agents believe that objects are in place when they are not.
This is in line with the finding of the paper that what limits the agents is recovering accurate positions from images.
The agents see what is in the scene, but the positions that they recover from images are not always accurate enough to act on.

\textbf{How accurate the placed objects are.}
The simulator's state shows how close the adaptive system comes when it works from sharp images.
In the six Mosaic tasks with 4 or 6 arms, we compare the final position of every cube with the center of its cell.
The median error of a task is 1.0 to 2.1\,mm, and 90\% of its cubes are within 2.3 to 4.5\,mm.
Of the 702 cubes, 572 lie within 3\,mm, 128 lie between 3 and 6\,mm, and 2 lie beyond 6\,mm.
These two are the faulty cubes of the uncompleted task with 6 arms.
This error contains both the error of the position that the agents measured in the images and the error of the motion that followed.

\begin{figure*}[t]
  \centering
  \includegraphics[width=\linewidth]{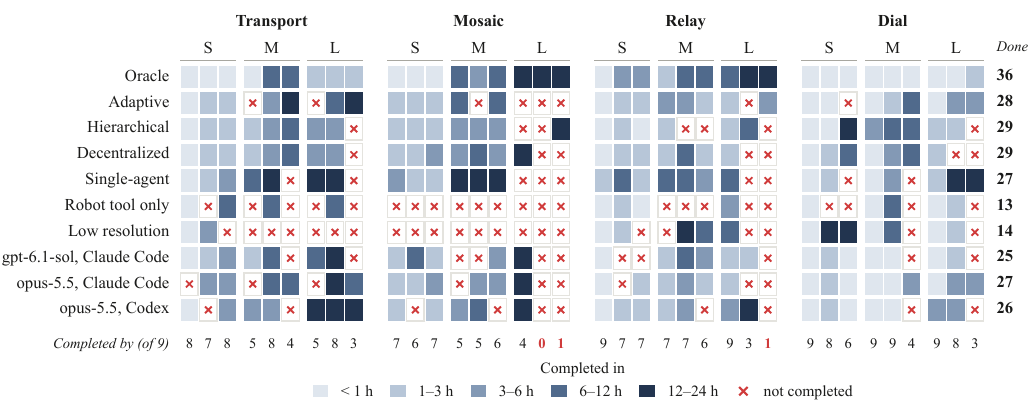}
  \caption{The outcome of every task under ten conditions. A darker cell means that the task took longer, and a cross means that it was not completed. S, M, and L are the team sizes, and the three cells within each are Levels 1 to 3. The bottom row counts the conditions other than the Oracle that complete a task, and the right column the tasks that a condition completes.}
  \label{fig:grid}
\end{figure*}

\textbf{Repeated runs.}
We ran the adaptive system twice more on the 12 Level 3 tasks.
The three runs complete 10, 8, and 9 of them.
\Cref{tab:repeatfailures} lists the seven failures of the two repeated runs.
No task fails in all three runs, and six are completed in all three.
The failures look like those of the first run.
Objects fall out of reach, cubes end a few millimeters outside the tolerance, and the largest Mosaic task slows as the picture fills.
In Mosaic with 8 arms, 7 of the 15 off-center cubes miss the 6\,mm limit by less than 2\,mm.
The Dial task with 8 arms fails in two of the three runs, both times after a module falls from the table.
In run 3, the agents stopped the Dial task with 16 arms at 11 hours, after long and repeated failure to recover a module that had been flung off the table.

\section{What is lost at low resolution}
\label{sec:supp-lowres}

With low resolution, \mbox{gpt-6.1-sol} completes only 14 tasks (\cref{tab:allmetrics}).
In their skill files, its agents write down what they can no longer do when every image has only $128\times80$ pixels.
\Cref{tab:lowres} quotes a few of these notes, one for each visual ability that the images no longer support.

At low resolution, a camera still draws its image at the normal size.
Blocks of pixels are then averaged into single pixels, which gives $128\times80$.
The cameras and the camera data in the task file stay the same.

\begin{table}[t]
  \caption{What agents say they can no longer do at $128\times80$ pixels. Each quote comes from the skill file of one run at low resolution.}
  \label{tab:lowres}
  \centering
  \footnotesize
  \setlength{\tabcolsep}{4pt}
  \begin{tabular}{@{}>{\raggedright\arraybackslash}p{0.32\linewidth}>{\raggedright\arraybackslash}p{\dimexpr0.68\linewidth-2\tabcolsep\relax}@{}}
    \toprule
    Ability & Note of the agents \\
    \midrule
    Position to millimeters & ``more than twice the 6mm placement tolerance'' (Mosaic) \\
    Separating objects & ``Adjacent same-color cubes can merge at 128x80'' (Mosaic) \\
    Reading symbols & ``tiny glyphs and three-label planar-pose ambiguity'' (Dial) \\
    Depth from views & ``roughly one-pixel center change moved inferred disk depth by 110mm'' (Dial) \\
    Fine seating & ``Low resolution does not directly resolve blade-depth tolerances'' (Transport) \\
    \bottomrule
  \end{tabular}

\end{table}

The agents respond with more computation on the images.
For example, they write code that measures an image, such as fitting the known size of a cube to its outline.
They also combine several camera views, and they enlarge or crop an image in code, although one note observes that an enlargement holds no extra image resolution.

The agents also look more often.
In the Level 3 runs at low resolution, they look at the scene again after a failed action, before the next action, in 28\% of the cases.
In the Level 3 runs with the robot tool only, they do so in 10\% of the cases.

\begin{figure*}[t]
  \centering
  \includegraphics[width=\linewidth]{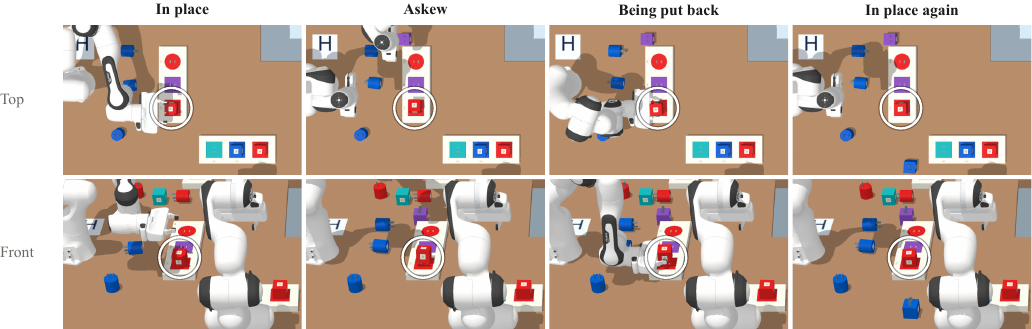}
  \caption{A delivered item that comes loose and is put back, in the run of \mbox{gpt-6.1-sol} with the adaptive system on Transport with 8 arms at Level 2. The circled cube is plugged into its socket, comes askew, is lifted, and is plugged in again. The rows are the top and front overviews, each cropped to the same region.}
  \label{fig:recovery}
\end{figure*}

\section{Additional results}
\label{sec:supp-results}

\textbf{All numbers.}
\Cref{tab:allmetrics} gives the time, tokens, messages, tool calls, and failure rates of the 14 conditions.
For the failure rates, it adds the spread over the 36 tasks and the number of failed grasps to what the paper shows.

\textbf{Outcome of every task.}
\Cref{fig:grid} gives the outcome of each of the 36 tasks under ten conditions.
These are the Oracle, the four agent systems, the two conditions that take away code tools or image resolution, the two best models in Claude Code, and \mbox{opus-5.5} in Codex.
In Transport with 8 and 12 arms, fewer conditions complete Level 1 than Level 2 (5 against 8 of 9).
Once an item is plugged into a socket, the socket holds it in place, whereas a flat target needs the item placed inside its square by position from images alone.

\begin{figure}[t]
\setlength{\fboxsep}{5pt}%
\begin{lrbox}{\suppcodebox}\begin{minipage}{\dimexpr\linewidth-2\fboxsep-2\fboxrule\relax}
\fontsize{6.2}{7.4}\selectfont
\begin{verbatim}
from PIL import Image
import numpy as np, math, sys

def centroid(path, roi):
    a = np.array(Image.open(path))
    x0, x1, y0, y1 = roi; b = a[y0:y1, x0:x1]
    m = (b.min(2) > 150) & ((b.max(2) - b.min(2)) < 65)
    ys, xs = np.where(m)
    return [xs.mean() + x0, ys.mean() + y0]

def ray(c, q, uv, fov):
    w, x, y, z = q
    R = np.array(
        [[1-2*(y*y+z*z), 2*(x*y-z*w), 2*(x*z+y*w)],
         [2*(x*y+z*w), 1-2*(x*x+z*z), 2*(y*z-x*w)],
         [2*(x*z-y*w), 2*(y*z+x*w), 1-2*(x*x+y*y)]])
    f = 400 / math.tan(math.radians(fov / 2)); u, v = uv
    d = R @ np.array([(u - 639.5) / f,
                      -(v - 399.5) / f, -1])
    return np.array(c), d / np.linalg.norm(d)

def tri(uv1, uv2):
    c1, d1 = ray([-.481, 0, 1.65],
                 [.6576, .2600, .2600, .6576], uv1, 50)
    c2, d2 = ray([-.681, -.4, 1.42],
                 [.8672, .3392, .1328, .3396], uv2, 38)
    A = np.column_stack([d1, -d2])
    t = np.linalg.lstsq(A, c2 - c1, rcond=None)[0]
    p1 = c1 + t[0] * d1; p2 = c2 + t[1] * d2
    return (p1 + p2) / 2, np.linalg.norm(p1 - p2)

if __name__ == '__main__':
    uv1 = centroid(sys.argv[1],
                   list(map(int, sys.argv[3:7])))
    uv2 = centroid(sys.argv[2],
                   list(map(int, sys.argv[7:11])))
    print(uv1, uv2, tri(uv1, uv2))
\end{verbatim}
\end{minipage}\end{lrbox}
\noindent\fcolorbox{skillrule}{skillbg}{\usebox{\suppcodebox}}
  \caption{A script that a delegate wrote during a Transport run, reformatted. The agent typed the data of the two cameras into the script.}
  \label{fig:code}
\end{figure}

\textbf{A delivered item that comes loose.}
A task is often completed only after things go wrong on the way (main paper, Sec.~4.3).
\Cref{fig:recovery} shows a case in the run of \mbox{gpt-6.1-sol} with the adaptive system on Transport with 8 arms at Level 2.
An arm plugs a cube into its socket and lets go.
Later, while the arm at that board moves, the cube comes out of place: it sits 12\,mm off and tilted by 17 degrees.
The agents see it, lift the cube, and try again.
The cube first comes to rest on the rim of the socket, and only a later try seats it.
In this run, an item that counted as delivered stopped counting 8 times.
In 5 of these cases it counted again within seconds, and in 3 it took longer; the figure shows the longest.

\begin{table}[t]
  \caption{Each condition compared with \mbox{gpt-6.1-sol} with the adaptive system over the same 36 tasks. The adaptive system completes 28 tasks, with a 95\% interval of 23 to 32. The two middle columns count the tasks that only one of the two completes. The last column is the $p$-value of an exact McNemar test after Holm's correction.}
  \label{tab:paired}
  \centering
  \footnotesize
  \setlength{\tabcolsep}{3.5pt}
  \begin{tabular*}{\linewidth}{@{\extracolsep{\fill}}lcccc@{}}
    \toprule
     & Tasks & Only & Only this & \\
     & completed & Adaptive & condition & $p$ \\
    \midrule
    Oracle & 36 & 0 & 8 & 0.055 \\
    \midrule
    Hierarchical & 29 (24--33) & 5 & 6 & 1 \\
    Decentralized & 29 (24--33) & 4 & 5 & 1 \\
    Single-agent & 27 (22--32) & 4 & 3 & 1 \\
    \midrule
    Robot tool only & 13 (8--18) & 15 & 0 & $<$0.001 \\
    Low resolution & 14 (10--18) & 15 & 1 & 0.004 \\
    \midrule
    \mbox{gpt-6.1-sol}, Claude Code & 25 (20--30) & 8 & 5 & 1 \\
    \mbox{opus-5.5}, Claude Code & 27 (22--32) & 4 & 3 & 1 \\
    \mbox{opus-5.5}, Codex & 26 (21--31) & 8 & 6 & 1 \\
    \bottomrule
  \end{tabular*}

\end{table}

\begin{table}[t]
  \caption{Median time per task at each team size, with 12 tasks per size. The number of completed tasks is in parentheses.}
  \label{tab:teamsize}
  \centering
  \footnotesize
  \setlength{\tabcolsep}{6pt}
  \begin{tabular}{@{}lccc@{}}
    \toprule
     & Small & Medium & Large \\
    \midrule
    Adaptive & 1.4 h (11) & 4.4 h (10) & 9.2 h (7) \\
    Hierarchical & 1.2 h (12) & 3.9 h (10) & 5.5 h (7) \\
    Decentralized & 1.4 h (12) & 4.4 h (12) & 7.1 h (5) \\
    Single-agent & 2.2 h (11) & 11.7 h (10) & 22.0 h (6) \\
    \bottomrule
  \end{tabular}

\end{table}

\textbf{Code that agents write.}
\Cref{fig:code} shows a script of the kind that the paper describes (main paper, Sec.~4.5).
A delegate wrote it in the run of \mbox{gpt-6.1-sol} with the adaptive system on Transport with 12 arms at Level 1, to locate a cylinder.
It finds the bright label on top of the cylinder in the images of two cameras and returns the point closest to the two viewing rays.
Eight seconds after its first result, the delegate moved an arm to a point above the cylinder.
Such a script does not make every position accurate: the same run ended with four cylinders 5.5 to 8.1\,mm off center (\cref{tab:failures}).

\begin{figure*}[t]
  \setlength{\suppmsgh}{88pt}%
  \begin{minipage}[t]{0.305\linewidth}
  {\footnotesize\textbf{Hierarchical}}\par\vspace{2pt}
  \skillboxh{\suppmsgh}{\textbf{Worker C to the controller:} C white overhead approach failed timeout [...] Request D move/rotate retreat north to clear own south-stock approach [...]\\[3pt]
  \textbf{Controller to C and D:} [...] C retreat to clear own sector first, report whole-arm clearance. D do NOT move into shared space until that report [...]}
  \end{minipage}\hfill
  \begin{minipage}[t]{0.305\linewidth}
  {\footnotesize\textbf{Decentralized}}\par\vspace{2pt}
  \skillboxh{\suppmsgh}{\textbf{Worker F to worker G:} Please retreat G full home to enable F slot1 alternate-yaw retry. [...] Need your actual clearing report before my next descent.\\[3pt]
  \textbf{G to F, 30 seconds later:} G FULL HOME joint retreat completed [...] Your alternate yaw local recovery may execute; no G approach meanwhile.}
  \end{minipage}\hfill
  \begin{minipage}[t]{0.37\linewidth}
  {\footnotesize\textbf{Adaptive}}\par\vspace{2pt}
  \skillboxh{\suppmsgh}{\textbf{Delegate to the controller:} [...] Candidate staging x$\sim$.65,y$\sim$.55 may overlap D hover [...]; can you keep D from entering north-central table and move its empty wrist home if feasible? [...]\\[3pt]
  \textbf{Controller to the delegate:} D now home empty [...] D will remain home through your E recovery; you may use x$\sim$.65,y$\sim$.55 stage with local clearance assessment. [...]}
  \end{minipage}
  \caption{One need under three agent systems, in the example runs of \cref{tab:firsthour}: an agent asks that another arm move out of its way. The messages are quoted as sent, and [...] marks what is left out.}
  \label{fig:messages}
\end{figure*}

\textbf{Paired tests.}
Every condition runs the same 36 tasks, so we can compare a condition with the adaptive system task by task (\cref{tab:paired}).
We do this for the nine other conditions of the two best models.
We leave out the four other models, which complete at most 3 tasks.
For each condition, we count the tasks that only the adaptive system completes and the tasks that only the other condition completes.
An exact McNemar test on these two counts gives a $p$-value, and we correct the nine $p$-values with Holm's method.
The intervals come from 20{,}000 bootstrap resamples of the tasks, drawn within each task type.

Two conditions differ from the adaptive system: the robot tool only and low resolution are worse.
The Oracle completes 8 more tasks and none fewer, but with nine tests this is not significant after correction ($p=0.008$ before correction and $0.055$ after).
The four combinations of the two best models and the two harnesses do not differ detectably from each other: the smallest of their six $p$-values is 0.58 before correction.
Among the five conditions of \mbox{gpt-6.1-sol} that keep sharp images and code tools, no pair differs detectably: the smallest of the ten $p$-values is 0.29 before correction.

\begin{table}[t]
  \caption{The first hour of one example run of each agent system on one task, Transport with 8 arms at Level 2, with \mbox{gpt-6.1-sol}.}
  \label{tab:firsthour}
  \centering
  \footnotesize
  \setlength{\tabcolsep}{4pt}
  \begin{tabular*}{\linewidth}{@{\extracolsep{\fill}}lcccc@{}}
    \toprule
     & Agents at the & Agents over & & Pairs of sender \\
     & same time & the hour & Messages & and receiver \\
    \midrule
    Single-agent & 1 & 1 & 0 & 0 \\
    Hierarchical & 9 & 9 & 213 & 16 \\
    Decentralized & 8 & 8 & 174 & 49 \\
    Adaptive & 1--4 & 16 & 86 & 28 \\
    \bottomrule
  \end{tabular*}
\end{table}

The same resampling gives intervals for the speed of the multi-agent systems.
On the tasks that both complete, the adaptive, decentralized, and hierarchical systems are 2.3, 2.4, and 2.6 times as fast as the single agent in the median task.
The 95\% intervals are 1.9 to 2.5, 1.8 to 2.7, and 2.3 to 3.4.

These tests treat the tasks as the sample.
Each task ran once per condition, so they do not measure how much a second run of the same task would differ.
The repeated runs of the adaptive system at Level 3 (\cref{sec:supp-failures}) give a first estimate: across three runs, it completes 8 to 10 of the 12 tasks.

\textbf{Time by team size.}
\Cref{tab:teamsize} gives the median time per task of the four agent systems at each team size, with \mbox{gpt-6.1-sol}.
The single agent falls behind most on the largest teams.

\section{The four agent systems on one task}
\label{sec:supp-systems}

The paper compares the four agent systems by their results over all tasks.
Here we show how they work in one example run each, on one task, Transport with 8 arms at Level 2, with \mbox{gpt-6.1-sol}.

\begin{figure}[t]
  {\footnotesize\textbf{Scope that the controller gives a delegate with arms B and C}}\par\vspace{2pt}
  \skillbox{\textbf{goal:} Deliver one yellow source C cube into B\_slot2, seated released stable, using B/C and your choice of direct reach, safe staging or supported airborne handover.\\[2pt]
  \textbf{done\_when:} B\_slot2 visibly fully seated and released stable; B/C clear placement and transfer area, all calls resolved; promptly return.}

  \vspace{6pt}
  {\footnotesize\textbf{The delegate's last words}}\par\vspace{2pt}
  \skillbox{B\_slot2 seated, released, and stable; B\_slot1 preserved. B/C cleared, all actions resolved, control returned.}
  \caption{One delegation in the adaptive system. The upper box holds two fields of the scope that the controller writes for the delegate. The lower box holds the delegate's last words before its session ends.}
  \label{fig:delegation}
\end{figure}

\textbf{Who holds the arms and who talks.}
\Cref{tab:firsthour} counts the agents and their messages in the first hour of the task.
The single agent holds all 8 arms and sends no messages.
In the hierarchical system, 8 workers hold one arm each, and a controller holds none.
A worker cannot write to another worker, so every message passes through the controller.
This gives 16 pairs of sender and receiver: the controller with each worker, in both directions.
In the decentralized system, the same 8 workers write to each other directly, and 49 of the 56 possible pairs occur.
The adaptive system starts with one controller that holds every arm.
In the first hour it creates 15 delegates, never more than 3 at a time, and it sends fewer messages than the two fixed teams.

\begin{figure*}[t]
\begin{minipage}[t]{\columnwidth}
  {\footnotesize\textbf{Instruction to a delegate}}\par\vspace{2pt}
  \skillbox{\fontencoding{T1}\selectfont Your current runtime role is \{\{ACTOR\}\}, a temporary delegate, not a standing per-robot Worker. This runtime assignment replaces the inherited parent role, connection and team finish authority; the task, rules and evidence restrictions remain in force. Inherited pending calls and plans are history, not work to replay.\\[3pt]
  Assigned scope, as submitted by the Controller: \{\{SCOPE\}\}\\[3pt]
  Operate only your assigned robots, and only through your robot tool. The team has started: carry out the scope autonomously, handle parent messages and report milestones or blockers through the role protocol. When the scope is done, call return\_task with outcome and evidence, then end your turn; follow guide/adaptive.md for waiting, pausing and partial handback, and reread it after a context gap. Reread the headings of the run's skills.md before acting; teammates may have added entries since you were forked. Do not create delegates or other sessions, use administrative operations, modify the harness, read private evaluation data or finish the team run.}
  \caption{The instruction that a delegate of the adaptive system receives when it is created, as in the code that we will release. The runtime fills in the two fields in double braces. \texttt{guide/adaptive.md} is a part of the general guide, and \texttt{skills.md} is the skill file.}
  \label{fig:instructions}
\end{minipage}\hfill
\begin{minipage}[t]{\columnwidth}
  {\footnotesize\textbf{Format of an entry}}\par\vspace{2pt}
  \skillbox{\textbf{[fix $|$ method $|$ finding $|$ special-case $|$ other]} conclusion, with when it applies --- role --- tags\\[2pt]
  \textbf{Content:} what happened or what to do, with the values that matter.\\
  \textbf{Scope:} when it holds, and when it does not.\\
  \textbf{Evidence:} call IDs, images, measured values.\\
  \textbf{Confidence:} high, medium, or low, and why.}

  \vspace{6pt}
  {\footnotesize\textbf{An entry written by an agent}}\par\vspace{2pt}
  \skillbox{\textbf{[fix] Cast RGB arrays to float before integer-ratio color masks --- Controller --- tags: grasp}\\[2pt]
  \textbf{Problem:} A scratch OpenCV red-pixel mask used \texttt{red > 2*green} on uint8 channels. Integer multiplication wrapped, admitting non-red table/robot pixels and corrupting fitted timber lines.\\
  \textbf{Content:} Convert the image to float before channel-ratio multiplication, or use floating multipliers. [\ldots] Do not reuse geometry derived from the wrapped mask.\\
  \textbf{Scope:} Local RGB analysis only; it does not correct occlusion, camera rounding or changes in payload pose.\\
  \textbf{Confidence:} high; arithmetic issue directly reproduced.}
  \caption{The skill file. The first box gives the format that the general guide asks for. The second box is an entry that a controller wrote in a Relay run at low resolution. It records a mistake in the agent's own image analysis and the fix.}
  \label{fig:skillentry}
\end{minipage}
\end{figure*}

\textbf{Asking another arm to move.}
\Cref{fig:messages} shows one need under the three multi-agent systems: an agent wants another arm out of its way.
The hierarchical worker asks the controller, which then instructs both workers.
The decentralized worker asks its neighbor, which answers 30 seconds later.
The adaptive delegate asks the controller, because the controller holds that arm at the time.
In this run, delegates never write to each other, and the controller passes on what they need.

\textbf{One delegation.}
\Cref{fig:delegation} shows the start and the end of one delegation.
The controller hands arms B and C to a delegate, with a goal and a condition for being done.
The delegate starts 24 robot actions, exchanges 3 messages with the controller, and returns the arms with a report after 7.8 minutes.
Over the 36 tasks, the controller of the adaptive system delegates between 0 and 129 times per task.
The median is 6 in Dial, 10 in Mosaic and in Relay, and 22 in Transport.

\begin{table}[t]
  \caption{Skill files of the two best models with the adaptive system, \mbox{gpt-6.1-sol} in Codex and \mbox{opus-5.5} in Claude Code, over 36 runs each. The values per run and per hour are medians over the runs.}
  \label{tab:skills}
  \centering
  \footnotesize
  \setlength{\tabcolsep}{5pt}
  \begin{tabular}{@{}lcc@{}}
    \toprule
     & \mbox{gpt-6.1-sol} & \mbox{opus-5.5} \\
    \midrule
    Runs with entries & 36 & 31 \\
    Entries per run & 11 & 6 \\
    Revisions per run & 11 & 0 \\
    Words per run & 4{,}130 & 1{,}395 \\
    Words per hour of work & 1{,}097 & 515 \\
    \midrule
    All entries & 541 & 433 \\
    All revisions & 971 & 6 \\
    \bottomrule
  \end{tabular}

\end{table}

\section{Written material and the skill file}
\label{sec:supp-start}

Every agent starts with the same written material.
It gets the rules of the benchmark, which say what it may read and use and how a run ends.
It gets a short instruction for its role, such as controller, worker, or delegate.
It gets a general guide to working with the arms and with other agents, together with the reference of the robot tool.
It gets a guide for the task type, for example on placing and releasing cubes in Mosaic.
It also gets the task file.
None of this material holds object positions.

\Cref{fig:instructions} gives one of these instructions in full, the one that a delegate of the adaptive system receives when it is created.
The runtime fills in the name of the delegate and the scope that the controller wrote, as in \cref{fig:delegation}.

\textbf{The skill file.}
The skill file is separate from these guides.
It starts empty in every run, and only the agents write to it.
The general guide asks for one format (\cref{fig:skillentry}).
The heading of an entry states its conclusion, so that a reader can tell from the heading alone whether the entry matters.
The body gives the content, the conditions under which it holds, the evidence, and the confidence.
An agent never changes an earlier entry.
It adds a revision below it and says what the revision changes.
The second box of \cref{fig:skillentry} shows one entry as an agent wrote it.

\begin{table}[t]
  \caption{The operations of the robot tool. The last two rows exist in one condition each: with the robot tool only, and in the Oracle.}
  \label{tab:tool}
  \centering
  \footnotesize
  \setlength{\tabcolsep}{4pt}
  \begin{tabular}{@{}>{\raggedright\arraybackslash}p{0.28\linewidth}>{\raggedright\arraybackslash}p{\dimexpr0.72\linewidth-2\tabcolsep\relax}@{}}
    \toprule
    Operation & What it does \\
    \midrule
    \texttt{ready}, \texttt{robot\_model} & Return the arms that an agent controls, the limits on its commands, and the geometry and limits of the arm \\
    \texttt{observe} & Returns RGB images from the named cameras and the state of the arms \\
    \texttt{public\_image} & Returns a picture listed in the task material, such as the reference picture in Mosaic \\
    \texttt{move} & Moves the tool of an arm along a straight line to a position and orientation, or moves its joints to given angles \\
    \texttt{gripper} & Opens, closes, or sets the opening of a gripper. In its grasp mode, it closes on an object of an expected width with a chosen force \\
    \texttt{action\_batch} & Sends several commands for several arms in one call \\
    \texttt{check\_motion} & Screens planned moves of several arms for overlap before they are sent \\
    \texttt{stop}, \texttt{recover\_action} & Brake a motion that is under way, and confirm that a motion has stopped \\
    \texttt{job}, \texttt{pending} & Report the status of an earlier command, and list the commands that are not yet resolved \\
    \texttt{wait\_events} & Waits for a message, an event, or the end of a command \\
    \texttt{send}, \texttt{inbox}, \texttt{history} & Exchange messages with other agents, and list the messages so far \\
    \texttt{work}, \texttt{checkpoint}, \texttt{yield} & Read and save an agent's own work note. \texttt{yield} saves it before the agent ends its turn \\
    \texttt{delegate}, \texttt{return\_task} & In the adaptive system, hand arms to a delegate and return them \\
    \texttt{team} & In the hierarchical system, shows the controller the state of its workers \\
    \texttt{finish} & Ends the run, when the agents declare the task done or stop it \\
    \midrule
    \texttt{public\_file}, \texttt{skills} & Read the task material, and read and extend the skill file \\
    \texttt{state} & Returns the position and velocity of every object and joint, and the progress that the evaluator tracks \\
    \bottomrule
  \end{tabular}
\end{table}

\textbf{Use of the skill file.}
Both of the best models use the skill file, but in different ways (\cref{tab:skills}).
\mbox{gpt-6.1-sol} writes entries in all 36 of its adaptive runs.
It writes about three times as many words per run as \mbox{opus-5.5}, and about twice as many per hour of work.
It also keeps its entries up to date.
Its 541 entries carry 971 revisions, and \mbox{opus-5.5} revises an entry only 6 times in all its runs.

This matches how each model works.
\mbox{gpt-6.1-sol} divides the work, and its delegates write 69\% of the entries and 74\% of the revisions.
The file is where one delegate leaves what the next one needs.
\mbox{opus-5.5} works mostly as one agent and writes shorter notes for itself.
In five runs, all of them Dial, it writes nothing.
It also follows the format less closely.
Of its entries, 15\% have no type, and others use types of its own, such as ``technique'' and ``pitfall''.
In both models, the most common tag of an entry is ``grasp''.

\section{The robot tool}
\label{sec:supp-tool}

Agents act through one tool with a small set of operations (\cref{tab:tool}).
Every motion command must state its speed and acceleration, and the tool accepts only values within the limits of the real Franka Emika Panda.
A command returns whether it completed or failed, with the reason.

The world does not wait for a command to end.
A motion command can return as soon as it is accepted, so that an agent can look, plan, or send messages while its arm moves.
The agent then asks for the outcome with \texttt{job} or waits for it with \texttt{wait\_events}.
Each arm carries out one command at a time, and an agent can brake its own motion with \texttt{stop}, which does not pause the world.

\begin{figure}[t]
  {\footnotesize\textbf{Call}}\par\vspace{2pt}
  \skillbox{\fontencoding{T1}\ttfamily\frenchspacing \{"op": "gripper", "data": \{"robot": "C", "mode": "grasp", "width\_m": 0.06, "force\_n": 45, "speed\_m\_s": 0.025\}, "views": ["C\_front", "C\_side"]\}}

  \vspace{6pt}
  {\footnotesize\textbf{Return (excerpt)}}\par\vspace{2pt}
  \skillbox{\fontencoding{T1}\ttfamily\frenchspacing "execution": \{"arm": "C", "type": "gripper", "status": "failed", "reason": "Grasp stopped outside expected object-width tolerance", [...]\}, "execution\_feedback": \{"actual\_width\_m": 0.0651, [...]\}}
  \caption{One call of the robot tool and a part of its return, in the run of \mbox{gpt-6.1-sol} with the adaptive system on Transport with 8 arms at Level 2. [...] marks what is left out.}
  \label{fig:toolcall}
\end{figure}

\begin{table}[t]
  \caption{Number of cameras in a task, by task type. The number grows with the team size and, in Dial, with the level.}
  \label{tab:cameras}
  \centering
  \footnotesize
  \setlength{\tabcolsep}{5pt}
  \begin{tabular}{@{}lcccc@{}}
    \toprule
     & All & At arms & Overviews & Close-ups \\
    \midrule
    Transport & 16--42 & 8--24 & 4 & 4--14 \\
    Mosaic & 14--24 & 8--16 & 4 & 2--4 \\
    Relay & 23--49 & 16--32 & 4 & 3--13 \\
    Dial & 29--69 & 8--32 & 4 & 17--33 \\
    \bottomrule
  \end{tabular}
\end{table}

An agent also keeps a short work note.
It saves the note with \texttt{checkpoint} or \texttt{yield} and reads it back with \texttt{work}, so that it can take up its work again after a pause.

\begin{figure*}[t]
  \centering
  \includegraphics[width=\linewidth]{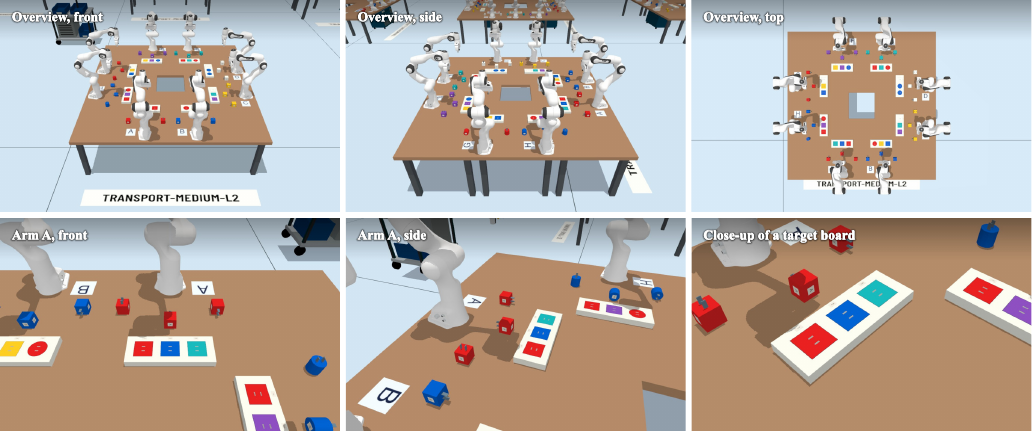}
  \caption{Images of six of the 30 cameras of one task at its start, Transport with 8 arms at Level 2, drawn by the renderer that serves the agents: the front, side, and top overviews, the two cameras at arm A, and the close-up of one target board.}
  \label{fig:cameraviews}
\end{figure*}

With the robot tool only, agents have no code tools, so they cannot read or write files.
Two further operations then let them read the task material and read and extend the skill file.
The Oracle gets no camera images.
Its \texttt{observe} returns only the state of the arms, and one further operation, \texttt{state}, returns the position and velocity of every object and joint, and the progress that the evaluator tracks.
In everything else, the Oracle uses the same tool with the same limits as the other conditions.

\Cref{fig:toolcall} shows one call and its return.
A delegate asks arm C to grasp an object that is 60\,mm wide.
The gripper stops at 65.1\,mm, so the tool reports that the grasp failed and gives the reason.
In the same run, moves failed for two other reasons: the requested pose could not be reached, or the motion did not end in time.

\section{Cameras}
\label{sec:supp-cameras}

\Cref{tab:cameras} gives the number of cameras in a task.
Each arm has one camera in front of it and one at its side.
Three of the four overview cameras look across the scene, and one looks down from above.
The other cameras are close-ups of the task objects, such as the target boards in Transport or the dials in Dial.
\Cref{fig:cameraviews} shows the images of six of the 30 cameras of one task at its start.
An agent asks for images by naming cameras, and it may name several at once.

\section{Aligning Codex and Claude Code}
\label{sec:supp-harness}

In both harnesses, a delegate of the adaptive system is a fork of the controller's session.
Its history ends at the controller's delegate call, and its first input is the delegate message.

\begin{table}[t]
  \caption{Settings of the two best models in their own harnesses.}
  \label{tab:settings}
  \centering
  \footnotesize
  \setlength{\tabcolsep}{4pt}
  \begin{tabular}{@{}lcc@{}}
    \toprule
     & \mbox{gpt-6.1-sol} & \mbox{opus-5.5} \\
    \midrule
    Harness & Codex 0.165.0 & Claude Code 2.1.316 \\
    Reasoning effort & high & high \\
    Context window (tokens) & 300{,}000 & 300{,}000 \\
    History is compacted at & 267{,}000 & 267{,}000 \\
    \bottomrule
  \end{tabular}
\end{table}

The two harnesses differ in how they create this fork.
Codex keeps one app-server process alive, which serves the robot tool and forks a session on request.
Claude Code receives the robot tool as an MCP server.
It creates a fork by resuming the parent session under a new session, so a fork exists only once its first turn runs.
A gate holds the fork's tool calls until the runtime has checked its copied history.
When a model runs in the harness of the other provider, \mbox{gpt-6.1-sol} in Claude Code or \mbox{opus-5.5} in Codex, a connector passes the model calls of the harness to the model's API.

The cross-harness runs test whether the harness explains the differences between the two best models (\cref{sec:modelresults}).

\Cref{tab:settings} lists the settings of the two best models in their own harnesses, the same in all 36 runs of each.
Each harness compacts an agent's history near the limit of the context window.
In both harnesses, web search and the harness's own multi-agent feature are switched off.

\section{Scope and cost of the experiments}
\label{sec:supp-scope}

\textbf{One model carries the comparisons.}
The experiments aim to learn how agents control many arms, not to rank models, so most runs use one model.
\mbox{gpt-6.1-sol} runs eight of the 14 conditions (288 runs of up to 24 hours, plus 24 repeated runs at Level 3), and every comparison of agent systems, tools, and image resolutions is made within it.

\textbf{One run per condition.}
A full condition takes hundreds of hours and billions of tokens, so each ran once apart from the Level 3 repeats of \cref{sec:supp-failures}, and we draw conclusions only from large differences.

\textbf{Cost.}
The 14 conditions used 302 billion input tokens and about 3{,}060 hours of runs in total (\cref{tab:allmetrics}), not counting the repeated runs.
The eight conditions of \mbox{gpt-6.1-sol} used 226 billion of these tokens.

\textbf{Four task types.}
We kept the set small because today's models need hours and billions of tokens for it, but new variants can be added in the scene files without changing the task logic.
As models improve, the same tasks can measure how cheaply they are completed, not only whether they are.

\section{Release}
\label{sec:supp-release}

To let others check and extend the experiments, we will release the simulation environment and the scene files of the 36 tasks, the robot tool, the runtime with its sandbox and the adapters for both harnesses, the instructions and guides that agents receive, the evaluation scripts, and the records of the runs, with messages, tool calls, images, trajectories, and videos.

\end{document}